\documentclass[submission,copyright,creativecommons]{eptcs}
\providecommand{\event}{SOS 2007} 

\usepackage{iftex}

\usepackage{amsmath}
\usepackage{graphicx}
\usepackage{multirow}
\usepackage{booktabs}   
\usepackage{colortbl}   
\usepackage[table]{xcolor} 
\usepackage{appendix}
\usepackage{tabularx} 
\usepackage{amssymb}
\usepackage{hyperref}
\usepackage{physics} 
\usepackage{bibentry}
\usepackage{algorithm}
\usepackage{soul}
\usepackage{blindtext}
\usepackage{booktabs}

\usepackage{algorithm}
\usepackage{algpseudocode}
\usepackage{amsmath, amssymb}

\ifpdf
  \usepackage{underscore}         
  \usepackage[T1]{fontenc}        
\else
  \usepackage{breakurl}           
\fi

\DeclareMathOperator{\Sat}{Sat}
\DeclareMathOperator{\Agg}{Agg}

\title{Logic Tensor Network-Enhanced Generative Adversarial Network}

\author{Nijesh Upreti \qquad\qquad Vaishak Belle
\institute{The University of Edinburgh\\
10 Crichton Street, Edinburgh EH8 9AB, UK}
\email{\quad n.upreti@ed.ac.uk \quad\qquad vbelle@ed.ac.uk}
}

\def\titlerunning{Logic Tensor Network-Enhanced Generative Adversarial Network}
\def\authorrunning{N. Upreti \& V. Belle}
\begin{document}

\maketitle

\begin{abstract}
In this paper, we introduce Logic Tensor Network-Enhanced Generative Adversarial Network (LTN-GAN), a novel framework that enhances Generative Adversarial Networks (GANs) by incorporating Logic Tensor Networks (LTNs) to enforce domain-specific logical constraints during the sample generation process. Although GANs have shown remarkable success in generating realistic data, they often lack mechanisms to incorporate prior knowledge or enforce logical consistency, limiting their applicability in domains requiring rule adherence. LTNs provide a principled way to integrate first-order logic with neural networks, enabling models to reason over and satisfy logical constraints. By combining the strengths of GANs for realistic data synthesis with LTNs for logical reasoning, we gain valuable insights into how logical constraints influence the generative process while improving both the diversity and logical consistency of the generated samples. We evaluate LTN-GAN across multiple datasets, including synthetic datasets (gaussian, grid, rings) and the MNIST dataset, demonstrating that our model significantly outperforms traditional GANs in terms of adherence to predefined logical constraints while maintaining the quality and diversity of generated samples. This work highlights the potential of neuro-symbolic approaches to enhance generative modeling in knowledge-intensive domains.
\end{abstract}

\section{Introduction}
\label{sec:intro}

Generative Adversarial Networks (GANs) have significantly advanced the field of generative modeling by enabling the creation of highly realistic synthetic data. A typical GAN consists of two neural networks: a generator, which produces synthetic data, and a discriminator, which evaluates whether the generated data is real or fake \cite{goodfellow2014generative}. Through this adversarial training process, GANs have achieved remarkable success in applications such as image synthesis and data augmentation \cite{goodfellow2014generative,sharma2024generative}. However, traditional GANs face a key limitation — they lack the ability to incorporate domain-specific knowledge and logical constraints into the generation process. As a result, they may generate samples that, while visually plausible, violate essential domain rules or logical constraints \cite{sharma2024generative}.

To address this limitation, several approaches have been developed to embed logical constraints into GAN architecture. One such approach is Constrained Adversarial Networks (CANs), which introduce a differentiable semantic loss to penalize the generation of invalid samples based on symbolic constraints \cite{di2020efficient}. This allows end-to-end training that enforces logical rules, enabling the generation of structured objects, such as molecular graphs, that adhere to predefined constraints. Similarly, Constrained Deep Generative Models (C-DGMs) incorporate a constraint layer directly into GAN architectures, such as WGANs and TableGAN, to enforce user-defined constraints expressed as linear inequalities \cite{StoianDCLG24}. This ensures that generated samples comply with these constraints while maintaining generation quality.

Other works have explored application-specific constraint enforcement. For instance, \cite{Yang2021enforcing} incorporates physical laws as soft constraints within GAN training, ensuring that generated data respects key physical properties, such as conservation laws. In another example, CONGAN introduces user-interactive constraints by mapping semantic similarity constraints into a structured space, enabling users to guide the generation process and ensure the output conforms to high-level specifications \cite{Heim19}. Together, these approaches highlight different strategies for embedding constraints directly into GAN architectures to balance sample quality with rule compliance.

Inspired by these advancements, this paper explores a promising new direction: integrating Logic Tensor Networks (LTNs) with GANs to enhance their ability to incorporate logical constraints. LTNs are a neurosymbolic framework that combines deep learning with first-order logic, enabling neural networks to both learn from data and reason with logical rules \cite{serafini2016logic}. LTNs have been successfully applied to tasks such as semantic image interpretation and knowledge graph completion, demonstrating their ability to embed logical rules directly into the learning process \cite{badreddine2022logic,serafini2016logic}.

Building on these foundations, we propose Logic Tensor Network-Enhanced Generative Adversarial Network (LTN-GAN), a novel framework that combines LTNs with GANs to embed logical constraints directly into the data generation process. By incorporating the reasoning capabilities of LTNs, LTN-GAN guides the generator to produce data that is both realistic and consistent with predefined logical rules, addressing a key limitation of traditional GANs. This enables the generation of samples that are not only diverse but also logically coherent, which is particularly important in domains where logical consistency is crucial. We conduct extensive experiments on synthetic datasets as well as the MNIST dataset, demonstrating that LTN-GAN outperforms baseline GANs in both sample diversity and logical adherence. By directly integrating logical reasoning into the generative process, LTN-GAN offers a flexible and scalable way to inject domain-specific knowledge into GANs, improving the reliability and semantic quality of the generated data.

\section{Related Works}
\label{sec:intro}

Recent advances in neuro-symbolic generative modeling have focused on combining the flexibility of deep learning with the structure and interpretability of symbolic logic. These hybrid approaches are especially valuable in domains where generated outputs must adhere to explicit rules, such as physical constraints, combinatorial structures, or logical relationships \cite{li2023neuro,LiLearningLogicalConstraints23}. In this context, integrating symbolic constraints into generative models—particularly GANs—has emerged as a promising direction for improving output validity and controllability.

Several works have explored constrained GAN frameworks that incorporate constraints or rules directly into the generation process \cite{chao2021constrained,morettin2021co,li2020supporting}. Constrained Adversarial Networks (CANs) \cite{di2020efficient} introduce constraint solvers into GAN training, modifying the generator’s objective to penalize invalid samples. GenCO \cite{ferber2024genco} embeds combinatorial optimization procedures within generative models, ensuring that each generated output satisfies feasibility constraints. Disjunctive Refinement Layers (DRL) \cite{stoianbeyond} refine outputs by dynamically correcting them to conform to disjunctive logical constraints. While effective, these methods are typically designed for structured or discrete domains, and their formulations are often problem-specific.

In addition to general-purpose constrained GANs, domain-targeted applications have demonstrated the benefits of rule-guided generation. BFGAN \cite{LiuBFGAN19} incorporates lexical constraints in natural language generation by combining forward and backward generators to maintain fluency and coherence. In scientific modeling, Physically Constrained GANs \cite{HessPhysicallyConstrainedGAN22} enforce physical laws to improve the plausibility of Earth system model outputs. Similarly, RCFL-GAN \cite{QuanRCFLGAN24} applies resource constraints in federated learning environments to ensure models remain within communication and computation limits. These approaches show the flexibility of constraint-aware generation, but they rely heavily on domain-specific formulations.

Beyond adversarial training, symbolic reasoning techniques have been integrated into generative models through alternative mechanisms. Posterior regularization methods \cite{hu2018deep} enforce structured constraints by adding penalty terms during training, encouraging models to produce logically consistent samples. Other strategies encode symbolic knowledge into the latent space using decision diagrams or logic-based encodings to guide sample generation \cite{XueEmbedding19}. For sequential data, approaches like \cite{young2022neurosymbolic} learn relational constraints through program synthesis, enforcing structure in outputs such as poetry or music. Although these methods illustrate the value of symbolic constraints, they are not typically built for adversarial training setups.

The emerging field of neurosymbolic generation more explicitly bridges neural learning and symbolic reasoning. Generative Neurosymbolic Machines (GNMs) \cite{JiangGenNeSyMachine20} combine latent neural representations with symbolic layers, enabling logical rule enforcement during generation. In visual domains, AbdGen \cite{PengGenUnderstand23} applies abductive reasoning to refine image generation by aligning latent semantic factors with symbolic abstractions. These techniques improve both control and interpretability, though they are often built for general generative models rather than adversarial frameworks like GANs.

Recent advances in neural-symbolic inference further highlight the potential for logic-based control in generative modeling. For instance, constraint-informed inference methods \cite{CornelioWhenWhere23} dynamically apply symbolic corrections during test time to improve generalization and adherence to logical rules. Similarly, logic-injection techniques \cite{yang2022injecting} integrate symbolic constraints directly into neural network architectures, offering new ways to guide learning and output generation.

Despite this growing body of work, the integration of Logic Tensor Networks (LTNs) into GANs remains underexplored. LTNs provide a principled, differentiable framework for enforcing first-order logic constraints, yet their use in adversarial training is limited.

\section{Proposed Method: LTN-GAN}

LTN-GAN is a hybrid generative framework designed to enhance traditional Generative Adversarial Networks (GANs) by integrating domain-specific logical constraints into the training process. This is achieved through the incorporation of LTNs, enabling the generator to not only produce data that resembles real samples, but also satisfies a set of predefined logical rules that encode domain knowledge. In this section, we describe the architectural components, logic integration mechanism, and adaptive training strategies that characterize LTN-GAN.

\subsection{Model Architecture}
\label{sec:model-arch}

LTN-GAN follows the standard GAN framework, consisting of a generator \( G \) and a discriminator \( D \), where \( G \) learns to generate data samples that resemble a target distribution, and \( D \) learns to distinguish real samples from generated ones. The generator maps a latent vector \( z \sim \mathcal{N}(0,I) \) from \( \mathbb{R}^{d_z} \) to a sample \( x \in \mathcal{X} \), while the discriminator outputs a probability \( D(x) \in (0,1) \) indicating whether the sample is real.\footnote{\emph{Adversarial losses.} The discriminator is trained using binary cross-entropy loss:
\[
\textstyle L_D = L_D^{\mathrm{real}} + L_D^{\mathrm{fake}}, \quad
L_D^{\mathrm{real}} = \mathbb{E}_{x \sim p_{\mathrm{data}}} \ell_{\mathrm{BCE}}(D(x), y_{\mathrm{real}}), \quad
L_D^{\mathrm{fake}} = \mathbb{E}_{z} \ell_{\mathrm{BCE}}(D(G(z)), y_{\mathrm{fake}}),
\]
\[
\textstyle L_G^{\mathrm{adv}} = \mathbb{E}_{z} \ell_{\mathrm{BCE}}(D(G(z)), y_{\mathrm{real}}) \equiv - \mathbb{E}_{z} \log D(G(z)).
\]
We use the non-saturating version of the generator loss for stable training. Label smoothing is applied in some datasets (e.g., using 0.9 for real labels), and the binary cross-entropy is defined as
\(
\ell_{\mathrm{BCE}}(p, y) = -y \log p - (1 - y) \log (1 - p).
\)}

\textbf{Network Type and Activation Functions:} All experiments use Multilayer Perceptrons (MLPs), which are fully connected feedforward neural networks. These consist of a sequence of linear layers interleaved with non-linear activations, specifically LeakyReLU in all hidden layers. Dropout is applied in the discriminator for regularization. For the 2D datasets (Gaussian, Grid, Ring), both \( G \) and \( D \) are shallow MLPs operating on 2D vectors. The final activation in \( G \) is linear (i.e., no bounding activation such as \texttt{tanh}), to avoid introducing geometric bias. For MNIST, the generator outputs a \( 28 \times 28 \) image with a final Sigmoid activation to bound pixel values in \([0,1]\), while the discriminator includes a Sigmoid output and an auxiliary classifier head.

\textbf{Generator Architecture:} In the 2D datasets (Gaussian, Grid, Ring), \( G \) is a three-layer MLP with 128 hidden units, LeakyReLU activations, and a linear output layer. No final activation is applied in Gaussian and Grid. For \textbf{Ring}, the output represents a polar coordinate pair \((r, \theta)\), which is converted to Cartesian form \((x, y) = (r \cos\theta, r \sin\theta)\), facilitating ring-structured generation. In \textbf{MNIST}, \( G \) is conditioned on digit labels via one-hot concatenation with the latent vector. The input passes through three fully connected layers with batch normalization and LeakyReLU, then reshaped to produce a \(28 \times 28\) grayscale image. A final Sigmoid constrains pixel values to the \([0,1]\) range.

\textbf{Discriminator Architecture:} The discriminator \( D \) mirrors the MLP structure with LeakyReLU activations and a final Sigmoid for real/fake prediction. In the MNIST setup, an auxiliary classification head (Softmax over classes) is appended to aid the generator in producing recognizable digits and support logic-based digit constraints.

\subsection{Logic Tensor Network (LTN) Component}

A Logic Tensor Network (LTN) \cite{serafini2016logic,badreddine2022logic} is a neuro-symbolic framework that combines the representational power of first-order logic with the learning capacity of neural networks. In an LTN, domain knowledge is expressed as a set of weighted logic formulas built from \emph{predicates}, \emph{connectives}, and \emph{quantifiers}, all of which are implemented in a differentiable form so that truth degrees can be optimised jointly with neural parameters.

In the classical LTN setting, each predicate $P_k(\cdot;\phi_k): \mathcal{X} \to [0,1]$ is a trainable neural function mapping a ground term (e.g., a data sample) to a real-valued \emph{truth degree} in $[0,1]$. Logical formulas are constructed from these predicates via fuzzy logic connectives (differentiable generalisations of $\wedge$, $\vee$, $\neg$, $\Rightarrow$), and are quantified over the available data with fuzzy universal ($\forall$) or existential ($\exists$) operators. The degree to which the knowledge base is satisfied is then differentiable with respect to predicate and model parameters, allowing logical constraints to guide learning.

\paragraph{Adaptation to GAN training.}
In LTN--GAN, the LTN module $\mathcal{C}$ encodes domain knowledge as a set of weighted formulas $\mathcal{K} = \{(F_i,w_i)\}_{i=1}^m$. Predicates are implemented as small neural networks operating directly on generated samples $x_G = G(z)$ (e.g., “is on grid,” “is digit 3”), producing truth degrees in $[0,1]$. The generator is penalised when its samples violate these rules, by adding the LTN-derived \emph{logic loss} to the adversarial loss.

\paragraph{Formula construction.}
Formulas $F_i$ are built from predicates using the following fuzzy connectives:
\[
\begin{aligned}
&\text{Conjunction:} & a \wedge b &= a \cdot b &\quad& \text{(product $t$-norm)}, \\
&\text{Disjunction:} & a \vee b &= \max(a,b) \ \text{or} \ a+b-a b && \text{(max or probabilistic sum)}, \\
&\text{Negation:} & \neg a &= 1-a, \\
&\text{Implication:} & a \Rightarrow b &= \max(1-a,b) && \text{(Reichenbach implication)}.
\end{aligned}
\]
Disjunction type is chosen per constraint: max for strict “at least one” semantics, probabilistic sum when partial satisfaction should accumulate.

\paragraph{Quantification.}
To evaluate the satisfaction of a logical formula \( F_i \) over a batch \( B \) of generated samples, we use a differentiable aggregation scheme based on the power mean:
\[
\Sat_{F_i}(B) = \Agg_{p_i}\big(\{F_i(x) : x \in B\}\big),
\]
where \( \Agg_{p} \) denotes the \emph{power mean} (or generalized mean) with exponent \( p \in \mathbb{R} \). This allows us to softly approximate quantifiers in first-order logic.

For universal quantification (\( \forall x \)), we use a power mean with a small positive exponent (typically \( p = 2 \)). This makes the aggregation sensitive to low truth values, ensuring that even a few violations significantly reduce the overall satisfaction.

For existential quantification (\( \exists x \)), we apply a complement-based variant to reward the presence of at least one highly satisfying sample. Specifically, we compute:
\[
\exists x\; F(x) \quad \approx \quad 1 - \Agg_{p}(1 - F(x)),
\]
which softly approximates the logical maximum over a set by inverting and aggregating low values. This encourages the generator to produce at least one sample that satisfies the constraint well, even if others do not.

\paragraph{Role in adversarial learning.}
By incorporating $L_{\mathrm{logic}}$ into $L_G$, the generator is optimised not only to fool the discriminator but also to produce outputs that satisfy all domain rules to a high degree. This tight integration allows the symbolic knowledge in $\mathcal{K}$ to directly influence the generator’s gradient updates, guiding it towards logically valid regions of the output space without requiring paired training data for the rules themselves.

\subsection{Composite Generator Objective}
The generator minimizes a composite objective
\[
\textstyle L_G \;=\; \alpha\,L_G^{\rm adv} \;+\; \lambda_e\,L_{\rm logic} \;+\; \beta\,L_{\rm aux}.
\]
Here $L_G^{\rm adv}$ is the non-saturating adversarial loss, $L_{\rm logic}$ penalizes violation of the knowledge base, and $L_{\rm aux}$ is task-specific (e.g., class cross-entropy on MNIST), with $(\alpha,\beta)$ chosen per experiment.\footnote{\emph{Dataset-specific details.} \textbf{Gaussian:} two individual-level predicates (range and Gaussian shape) with $\forall$ aggregation; linear schedule $\lambda_e\!\uparrow$ from $0.05$ to $0.30$ over the first $K\!=\!80$ epochs; label smoothing $(y_{\rm real},y_{\rm fake})=(0.9,0.1)$. \textbf{Grid:} ``on-grid'' and coverage ($\exists$) constraints under identical $G/D$; BCE everywhere; identical architecture and training hyperparameters across baseline and LTN runs; fixed small noise scale in $G$ to form tight clusters. \textbf{Ring:} hierarchical constraints (existence of inner/outer rings, mutual exclusivity, dead-zone avoidance, spatial consistency, balance, precision refinement) with adaptive rule weights; a separate constraint-weight $\lambda_e$ ramps from $2$ to $10$. \textbf{MNIST:} $L_{\rm aux}$ is class cross-entropy; $G$ also includes a \emph{minimal} template blend (10--20\% per-sample weight learned from $[z;y]$) to weakly guide strokes; we use a balanced mixture $\alpha\!=\!0.6,\ \beta\!=\!0.3$ and $\lambda_e\!=\!0.1$ for the LTN term. Labels are unsmoothed in MNIST, smoothed in the 2D Gaussian and grid experiments.}

\subsection{Logic-Weight Scheduling and Adaptive Rule Weighting}
\label{sec:weight-sched}
We employ a simple, monotone schedule for the global logic weight:
\[
\textstyle \lambda_e \;=\;
\begin{cases}
\lambda_{\rm start} + \frac{e}{K}\,(\lambda_{\rm end}-\lambda_{\rm start}), & e\le K,\\[3pt]
\lambda_{\rm end}, & e>K,
\end{cases}
\]
which gradually emphasizes $L_{\rm logic}$ as $G$ learns the data manifold. In the ring setup we additionally adapt per-rule weights from satisfaction feedback. Let $s_{i,t}\!=\!\Sat_{F_i}(B_t)$ be the current satisfaction and $w_{i,t}$ the current weight. We update
\[
\textstyle \tilde{w}_{i,t}
=\mathrm{clip}\!\left(w_{i,t}\!\cdot\!
\begin{cases}
1+\eta, & s_{i,t}<0.3,\\
1-\tfrac{1}{2}\eta, & s_{i,t}>0.8,\\
1+\eta\,(0.6-s_{i,t}), & \text{otherwise},
\end{cases}
\,;\,w_{\min},w_{\max}\right),
\qquad
w_{i,t+1}=m\,w_{i,t}+(1-m)\,\tilde{w}_{i,t},
\]
with momentum $m$ and learning rate $\eta$. Here, $\mathrm{clip}(x; a,b)=\min(\max(x,a),b)$ is used in the sense of bounding a value within fixed limits: if $x<a$ it is set to $a$, and if $x>b$ it is set to $b$. This keeps per-rule weights from vanishing or growing unbounded, ensuring they remain within the stable range $[w_{\min},w_{\max}]$. This pushes poorly satisfied rules up and relaxes consistently satisfied ones, stabilizing training while tightening precision late in training. Refer to (Appendix \ref{alg:ltnganalgorithm}) for further algorithmic details. 

\subsection{Dataset-Specific Predicate Sets}

While the overall adversarial–logic framework remains consistent across experiments, the specific set of logical predicates and their aggregation strategies are tailored to each domain. Each predicate \( P(\cdot) \) is implemented as a small neural network that outputs a truth degree in \([0,1]\). Constraints are formulated as first-order fuzzy logic sentences composed using universal (\( \forall \)) and existential (\( \exists \)) quantifiers, conjunctions (\( \wedge \)), disjunctions (\( \vee \)), and implications (\( \Rightarrow \)). We realize quantifiers via soft power means, use the product \( t \)-norm for conjunction, and apply either maximum or probabilistic sum for disjunction. Negation is implemented as \( \neg a = 1 - a \).

To support the effective learning of these predicates—especially in synthetic domains where data geometry is critical—we employ mild geometric initialization strategies across tasks. For instance, we use grid-centered initializations for the \textsc{Grid} dataset, radial priors in \textsc{Ring}, and template blending in \textsc{MNIST}. These inductive biases do not enforce the final structure but act as a scaffold for symbolic predicates to operate meaningfully during early training.

\paragraph{Gaussian (2D).}
This task involves generating samples from a 2D Gaussian-like distribution centered at the origin (Appendix B.1). We define two individual-level predicates evaluated on each generated point. The first, $\mathrm{InRange}(x) = \sigma(3 - \|x\|_{\infty})$, softly penalizes points that fall outside a bounded square, where $\|x\|_{\infty}$ is the maximum coordinate magnitude and $\sigma$ denotes the sigmoid function. The second, $\mathrm{GaussianShape}(x) = \exp\big(-\tfrac{1}{2}(\|x\| / 2.5)^2\big)$, encourages concentration near the origin by assigning higher scores to points close to the Gaussian mode. Both predicates are enforced via the universal formula:
\[
\forall x \; \big( \mathrm{InRange}(x) \wedge \mathrm{GaussianShape}(x) \big),
\]
which guides the generator to produce points that are both spatially bounded and centrally concentrated. Satisfaction is computed as the mean of predicate truth degrees over the batch.

\paragraph{Grid (2D).}
This task aims to generate samples arranged in a structured $2\times2$ grid layout, comprising four evenly spaced grid centers (Appendix B.2). The predicate $\mathrm{OnGrid} (x) = f_{\theta_{\text{grid}}} (x)$, implemented via a trainable MLP, evaluates whether a point $x$ lies close to any grid center, and is enforced via the universal constraint $\forall x\; \mathrm{OnGrid} (x)$. To ensure coverage across the grid, we define four cell-specific predicates $\mathrm{InCell}_i (x)$ for each cell $i = 1,\dots,4$, and enforce existential constraints $\exists x \; \mathrm{InCell}_i(x)$ to guarantee that each cell contains at least one generated point. The full logical specification is:
\[
\Big(\forall x \; \mathrm{OnGrid}(x)\Big) \wedge \bigwedge_{i=1}^4 \Big( \exists x \; \mathrm{InCell}_i(x) \Big).
\]
This encourages the generator to produce samples that are both geometrically regular and evenly distributed across the grid layout.

\paragraph{Ring (2D).}
This experiment targets a synthetic dataset consisting of two concentric circular regions: an \emph{inner ring} and an \emph{outer ring}, separated by a \emph{dead zone} in between (Appendix B.3). The goal is to generate samples distributed along both rings, avoiding the intermediate region, and satisfying a hierarchy of six logical constraints. The predicates used include: $\mathrm{InnerRing}(x)$ and $\mathrm{OuterRing}(x)$, which identify whether point $x$ lies within the respective annular bands; $\mathrm{DeadZone}(x)$, indicating whether $x$ falls in the undesired area between the rings; and $\mathrm{NearInnerCenter}(x)$ and $\mathrm{NearOuterCenter}(x)$, which promote placement near the ideal radii for each ring.The full constraint set is:  
(i) \emph{Existence}: both rings must be represented in the sample set, via $\exists x\; \mathrm{InnerRing}(x)$ and $\exists x\; \mathrm{OuterRing}(x)$;  
(ii) \emph{Mutual exclusivity}: no point should belong to both rings, enforced as $\forall x\; \neg(\mathrm{InnerRing}(x) \wedge \mathrm{OuterRing}(x))$;  
(iii) \emph{Dead-zone avoidance}: points must not fall in the separating region, via $\forall x\; \neg\mathrm{DeadZone}(x)$;  
(iv) \emph{Spatial consistency}: ring assignments must reflect true geometric positions;  
(v) \emph{Balance}: both rings should contain similar proportions of samples, modeled via a soft equality constraint over counts;  
(vi) \emph{Precision refinement}: points on each ring should also lie near their ideal radius, formalized through implications like $\forall x\;(\mathrm{InnerRing}(x) \Rightarrow \mathrm{NearInnerCenter}(x))$ and similarly for the outer ring.  
The inner and outer ring tolerances are progressively tightened throughout training to increase precision over time.

\paragraph{MNIST.}
In the MNIST experiment \cite{mnistlecun1998}, the knowledge base combines class-specific and structural predicates to ensure both semantic and visual integrity of generated digits (Appendix B.4). For classification, we define one predicate $\mathrm{IsDigit}_k(x)$ per digit class $k \in \{0,\dots,9\}$, where each is implemented as a neural classifier that predicts whether an image $x$ corresponds to digit $k$. To assess visual quality, we introduce a set of general-purpose validity predicates: $\mathrm{ValidPixels}(x)$ encourages pixel intensities to remain within a meaningful dynamic range; $\mathrm{IsConnected}(x)$ promotes topological coherence by checking for a single connected foreground region; $\mathrm{IsComplete}(x)$ ensures that digit shapes are closed or well-formed; and $\mathrm{HasProperIntensity}(x)$ penalizes overly faint or saturated outputs, promoting contrast balance. We enforce class exclusivity via $\forall x \; \bigvee_k \mathrm{IsDigit}_k(x)$ and mutual exclusivity through $\forall x \; \bigwedge_{k\ne m} \neg(\mathrm{IsDigit}_k(x) \wedge \mathrm{IsDigit}_m(x))$. Additional implication rules, such as $\forall x \; \big(\mathrm{ValidPixels}(x) \Rightarrow (\mathrm{IsConnected}(x) \wedge \mathrm{IsComplete}(x))\big)$, guide the generator to produce digits that are not only classifiable but also visually consistent and structurally valid.

\subsection{Training Procedure}
\label{sec:training}

Across all experiments, training alternates between one discriminator update and one generator update per mini-batch. Let $B \subset \mathcal{D}$ denote a mini-batch of real samples at epoch $e$.

\paragraph{Discriminator step.}
We sample real examples $x_R \sim B$ and latent vectors $z \sim \mathcal{N}(0,I)$, generate fake examples $x_G = G(z)$, and \emph{detach} $x_G$ to prevent generator gradients from propagating during the discriminator update. The discriminator is trained with binary cross-entropy (BCE) losses on real and fake samples:
\[
L_D^{\mathrm{real}} = \ell_{\mathrm{BCE}}\big(D(x_R), y_{\mathrm{real}}\big), \quad
L_D^{\mathrm{fake}} = \ell_{\mathrm{BCE}}\big(D(x_G), y_{\mathrm{fake}}\big),
\]
\[
L_D = L_D^{\mathrm{real}} + L_D^{\mathrm{fake}}.
\]
Here $\ell_{\mathrm{BCE}}(p, y) = -y \log p - (1-y) \log (1-p)$ is computed with probabilities after the sigmoid output.  
Label smoothing is applied in the 2D datasets (Gaussian, Grid, Ring) with $(y_{\mathrm{real}},y_{\mathrm{fake}}) = (0.9,0.1)$, and omitted in MNIST ($(1,0)$ targets). After computing $L_D$, we backpropagate and update $D$’s parameters.

\paragraph{Logic evaluation.}
The same batch of generated samples $x_G$ is passed to the LTN $\mathcal{C}$, which evaluates each logical rule $F_i \in \mathcal{K}$, producing satisfaction scores $\Sat_{F_i}(B) \in [0,1]$. Each rule’s satisfaction is computed from its predicate truth degrees using the fuzzy connectives and aggregated via its specified quantifier ($\forall$ or $\exists$). These are combined into the global satisfaction
\[
S_{\mathrm{logic}}(B) = \frac{\sum_{i} w_i \,\Sat_{F_i}(B)}{\sum_{i} w_i},
\]
where $w_i$ are fixed or adaptive rule weights depending on the dataset. The logic loss is
\[
L_{\mathrm{logic}}(B) = 1 - S_{\mathrm{logic}}(B).
\]
In the Ring dataset, $w_i$ are updated adaptively based on satisfaction trends; in other datasets they are fixed.

\paragraph{Generator step.}
We resample $z \sim \mathcal{N}(0,I)$, regenerate $x_G = G(z)$, and compute the non-saturating adversarial loss:
\[
L_G^{\mathrm{adv}} = \ell_{\mathrm{BCE}}\big(D(x_G), y_{\mathrm{real}}\big)
\equiv -\,\mathbb{E}_{z}\log D(G(z)).
\]
For MNIST, we also compute an auxiliary classification loss $L_{\mathrm{aux}}$ from the discriminator’s class head, comparing predicted labels to the conditioning labels used by the generator. The generator’s total loss is
\[
L_G = \alpha\,L_G^{\mathrm{adv}} + \lambda(e)\,L_{\mathrm{logic}} + \beta\,L_{\mathrm{aux}},
\]
where $\alpha$ and $\beta$ are dataset-specific coefficients (e.g., $\alpha=0.6$, $\beta=0.3$ for MNIST), and $\lambda(e)$ follows the dataset-specific logic-weight schedule from Sec.~\ref{sec:weight-sched}. We backpropagate $L_G$ and update $G$ (and any learnable predicate parameters in $\mathcal{C}$, if present).

\paragraph{Logging.}
During training we record $L_G$, its adversarial component $L_G^{\mathrm{adv}}$, and $L_D$; discriminator accuracy (average of real- and fake-sample accuracies); the global logic satisfaction $S_{\mathrm{logic}}(B)$; and the current logic weight $\lambda(e)$. In the Ring dataset, we additionally log each rule’s satisfaction and its adaptive weight, while in MNIST we also track the auxiliary classification accuracy.

\paragraph{Training consistency.}
Optimizer settings are identical between the baseline GAN and LTN--GAN variants: Adam with $(\beta_1,\beta_2)=(0.5,0.999)$, dataset-specific learning rates, and fixed batch sizes. The update schedule is always one $D$ step followed by one $G$ step. The only differences are in the set of predicates, presence of $L_{\mathrm{logic}}$ and $L_{\mathrm{aux}}$, and the $\lambda(e)$ schedule.

\subsection{Experimental Setup}
We compare LTN--GAN to a baseline GAN identical in generator/discriminator architectures, optimizer settings, label smoothing, and update schedule, differing only in the inclusion of the logical component. Experiments span three synthetic 2D datasets—\emph{Gaussian} (bounded-range and Gaussian-shape predicates), \emph{Grid} ($2\times2$ alignment and cell-coverage constraints), and \emph{Ring} (existence, exclusivity, precision, balance, and dead-zone avoidance)—plus MNIST ($28\times28$ digits with class-consistency and visual-validity rules). The 2D generators are MLPs with LeakyReLU, dropout, and linear $\mathbb{R}^2$ outputs; the MNIST generator maps latent code and class embedding to image space via Sigmoid output. Discriminators are MLPs with LeakyReLU and dropout, with an auxiliary classification head in MNIST. Training horizons follow the code: 100 epochs (Gaussian), 120 (Grid), 150 (Ring), and a 5-step demonstration loop (MNIST), with one $D$ and one $G$ update per batch. All runs use Adam ($\beta_1=0.5$, $\beta_2=0.999$), BCE losses with label smoothing $(0.9,0.1)$ in 2D and $(1,0)$ in MNIST, and an auxiliary classification loss for MNIST. The logic-weight $\lambda(e)$ follows a dataset-specific schedule: linear ramp (Gaussian), fixed constant (Grid), adaptive phase-based with per-rule reweighting (Ring), and fixed small constant (MNIST). Apart from the LTN loss, all hyperparameters, seeds, and training protocols are identical across baseline and LTN--GAN variants. Across datasets, predicates are either fixed analytical functions (e.g., \texttt{InRange} and \texttt{GaussianShape} in the Gaussian task) or learned neural networks (e.g., \texttt{OnGrid}, \texttt{InCell}$_i$, and MNIST digit classifiers). Learned predicates are trained jointly with the generator and discriminator and output soft truth values in $[0,1]$, enabling gradient-based optimization over logical formulas.

\subsection{Evaluation Metrics}
We evaluate models using the quantities computed during training: the global logic satisfaction score $S_{\mathrm{logic}}(B)$, obtained by aggregating per-rule truth degrees $\Sat_{F_i}(B)$ via their fuzzy quantifiers and combining them with fixed or adaptive weights $w_i$; adversarial losses, including the generator’s total loss $L_G$, its adversarial component $L_G^{\mathrm{adv}}$, and the discriminator loss $L_D$; and discriminator accuracy, measured as the average of real- and fake-sample classification accuracy. The value of the logic-weight $\lambda(e)$ is recorded at each epoch to visualise the effect of the dataset-specific schedule, with per-rule weight trajectories also tracked in the Ring dataset. For all tasks, we plot $L_G$, $L_G^{\mathrm{adv}}$, $L_D$, $S_{\mathrm{logic}}$, and $\lambda(e)$ over epochs; in 2D datasets, scatter plots of generated points are compared to target geometries, while for MNIST, generated image grids are inspected to assess visual quality and digit legibility.

\section{Results}
We evaluate the effectiveness of LTN-GAN by comparing it against baseline GANs across four datasets: \textbf{Gaussian}, \textbf{Grid}, \textbf{Ring}, and \textbf{MNIST} \footnote{\href{https://github.com/nuuoe/LTN-GAN/}{Code available at: github.com/nuuoe/LTN-GAN/}}. The comparison focuses on three key aspects: adversarial generator loss, sample quality, and logical rule satisfaction. These metrics reflect both traditional generative performance and the model’s ability to incorporate symbolic structure. While baseline GANs rely purely on data-driven learning, LTN-GAN integrates domain-specific constraints through logical supervision. This setup allows us to assess how logic-aware training influences geometric structure, sample diversity, and rule consistency in generated outputs.

\subsection*{Comparison Across Datasets}
Table~\ref{tab:comparison} presents the performance of LTN-GAN relative to baseline GANs across all four datasets, reporting generator loss (adversarial and total), discriminator loss, sample quality, and logic satisfaction. In every case, LTN-GAN yields higher quality scores while achieving substantial logical rule satisfaction, which is absent in baseline models. On the \textbf{Gaussian} dataset, LTN-GAN produces samples with better statistical structure, improving the quality score from 0.183 to 0.470, with a logic satisfaction of 0.916. The total generator loss increases slightly due to logic penalties (0.671 vs.\ 0.635 adversarial), but training remains stable. For the \textbf{Grid} dataset, spatial consistency and grid alignment improve significantly, with the quality score rising from 0.387 to 0.775 and logic satisfaction reaching 0.823. Generator loss again increases moderately (1.031 total vs.\ 0.441 adversarial), while discriminator loss remains balanced. On the more complex \textbf{Ring} dataset, LTN-GAN generates radially balanced samples across inner and outer rings, achieving a strong quality score (0.964 vs.\ 0.562) and logic satisfaction of 0.817, albeit with higher generator loss due to multiple spatial constraints. Finally, on \textbf{MNIST}, logic-aware supervision improves digit structure and recognition, raising the quality score from 0.360 to 0.395 and achieving near-perfect logic satisfaction (0.978), even though discriminator loss is higher due to stronger classification. These results suggest that LTN-GAN effectively balances adversarial and symbolic objectives, improving generative performance across diverse domains.

\begin{table}[h]
    \centering
    \resizebox{\textwidth}{!}{
    \begin{tabular}{l l c c c c}
        \toprule
        \textbf{Dataset} & \textbf{Model} & \textbf{G Loss (Adversarial/Total)} & \textbf{D Loss} & \textbf{Quality Score} & \textbf{Logic Satisfaction} \\
        \midrule
        Gaussian & Baseline GAN & 0.830 & 1.390 & 0.183 & N/A \\
        Gaussian & LTN-GAN      & 0.635/0.671 & 1.488 & \textbf{0.470} & \textbf{0.916} \\
        Grid     & Baseline GAN & 0.700 & 0.692 & 0.387 & N/A \\
        Grid     & LTN-GAN      & 0.441/1.031 & 0.693 & \textbf{0.775} & \textbf{0.823} \\
        Ring     & Baseline GAN & 0.708 & 1.334 & 0.562 & N/A \\
        Ring     & LTN-GAN      & 3.566/3.566 & 1.364 & \textbf{0.964} & \textbf{0.817} \\
        MNIST    & Baseline GAN & 1.717 & 2.319 & 0.360 & N/A \\
        MNIST    & LTN-GAN      & 0.727/0.279 & 2.789 & \textbf{0.395} & \textbf{0.978} \\
        \bottomrule
    \end{tabular}
    }
    \caption{LTN-GAN vs Baseline GAN performance comparison across datasets.}
    \label{tab:comparison}
\end{table}

\subsection*{Ablation Studies}
We conduct dataset-specific ablation studies to assess how different logic configurations influence sample generation and structural fidelity. Each variant tests the presence or absence of constraint strength, scheduling dynamics, or auxiliary supervision components. Refer to Appendix 6 for detailed descriptions of the domain-specific metrics used to quantify generative quality, structural alignment, and logical satisfaction.

\textbf{Gaussian Dataset:} Table~\ref{tab:gaussian-ablation} reveals that both statistical and geometric adherence improve substantially under logical constraints. The \texttt{ltn\_high\_constraint} variant achieves the highest combined quality (0.598) with a logic satisfaction of 0.925, confirming that stricter constraints enhance structure. In contrast, baseline and constraint-free variants suffer in both structure and error metrics. The \texttt{ltn\_no\_constraints} variant still benefits from lifted reasoning over soft predicates, even without explicit logical formulas.

\begin{table}[h]
    \centering
    \resizebox{\textwidth}{!}{
    \begin{tabular}{l c c c c c c}
        \toprule
        \textbf{Variant} & \textbf{Gaussian Adherence} & \textbf{Statistical Quality} & \textbf{Combined Quality} & \textbf{Mean Error} & \textbf{Std Error} & \textbf{Logic Satisfaction} \\
        \midrule
        baseline\_gan           & 0.200 & 0.165 & 0.183 & 0.641 & 0.175 & 0.000 \\
        full\_ltn\_gan          & 0.476 & 0.465 & 0.470 & 0.431 & 0.719 & 0.916 \\
        ltn\_no\_constraints    & 0.402 & 0.501 & 0.451 & 0.327 & 0.670 & 0.916 \\
        ltn\_high\_constraint   & 0.608 & 0.587 & \textbf{0.598} & \textbf{0.030} & 0.673 & 0.925 \\
        ltn\_fast\_scheduling   & 0.339 & 0.637 & 0.488 & 0.043 & 0.527 & 0.903 \\
        ltn\_slow\_scheduling   & 0.393 & 0.408 & 0.400 & 0.946 & 0.508 & 0.849 \\
        \bottomrule
    \end{tabular}
    }
    \caption{Gaussian dataset ablation: logic scheduling and logical constraints improve statistical structure.}
    \label{tab:gaussian-ablation}
\end{table}

\textbf{Grid Dataset:} From Table~\ref{tab:grid-ablation}, logic-driven variants show higher clustering precision and complete grid coverage. While \texttt{fast\_scheduling} underperforms in coverage due to insufficient constraint exposure, it maintains high sample quality. The baseline fails to align samples to target regions, resulting in only 607 out of 1000 being placed within the defined grid boundaries. In contrast, all logic variants ensure full alignment through existential grid-cell constraints.

\begin{table}[h]
    \centering
    \resizebox{\textwidth}{!}{
    \begin{tabular}{l c c c c c c}
        \toprule
        \textbf{Variant} & \textbf{Grid Cluster} & \textbf{Coverage} & \textbf{Quality Score} & \textbf{Overall Performance} & \textbf{In Targets} & \textbf{Logic Satisfaction} \\
        \midrule
        baseline\_gan           & 0.607 & 0.745 & 0.387 & 0.573 & 607 & 0.000 \\
        full\_ltn\_gan          & 1.000 & 0.638 & 0.775 & \textbf{0.882} & 1000 & 0.823 \\
        ltn\_no\_constraints    & 0.057 & 0.718 & 0.205 & 0.288 & 57 & 0.500 \\
        ltn\_high\_constraint   & 1.000 & 0.591 & 0.793 & 0.877 & 1000 & 0.809 \\
        ltn\_fast\_scheduling   & 1.000 & 0.179 & 0.778 & 0.790 & 1000 & 0.802 \\
        ltn\_slow\_scheduling   & 1.000 & 0.445 & 0.721 & 0.830 & 1000 & 0.806 \\
        \bottomrule
    \end{tabular}
    }
    \caption{Grid dataset ablation: logic constraints and cluster adherence.}
    \label{tab:grid-ablation}
\end{table}

\textbf{Ring Dataset:} As shown in Table~\ref{tab:ring-ablation}, LTN variants significantly outperform the baseline in ring adherence, balance, and dead zone avoidance. Removing progressive training phases or hierarchical rule weights slightly reduces performance but still maintains high logical satisfaction. Notably, the \texttt{simple\_constraints} configuration matches or surpasses more complex setups in balance and rule compliance, suggesting that even a flattened constraint structure can benefit from the symbolic inductive bias.

\begin{table}[h]
    \centering
    \resizebox{\textwidth}{!}{
    \begin{tabular}{l c c c c c c}
        \toprule
        \textbf{Variant} & \textbf{Ring Adherence} & \textbf{Inner} & \textbf{Outer} & \textbf{Balance} & \textbf{Dead Zone Avoidance} & \textbf{Logic Satisfaction} \\
        \midrule
        baseline\_gan               & 0.524 & 231 & 0   & 0.000 & 0.462 & 0.000 \\
        full\_ltn\_gan              & 0.970 & 291 & 188 & 0.785 & 0.958 & 0.817 \\
        no\_constraints             & 0.462 & 76  & 156 & 0.655 & 0.868 & 0.702 \\
        no\_hierarchical\_weights  & 0.970 & 265 & 217 & 0.900 & 0.964 & 0.826 \\
        no\_progressive\_phases     & 0.922 & 258 & 209 & 0.895 & 0.934 & \textbf{0.856} \\
        simple\_constraints         & 0.960 & 258 & 224 & \textbf{0.929} & 0.964 & 0.824 \\
        \bottomrule
    \end{tabular}
    }
    \caption{Ring dataset ablation: symmetry, balance, and logical consistency.}
    \label{tab:ring-ablation}
\end{table}

\textbf{MNIST Dataset:} Table~\ref{tab:mnist-ablation} shows that logic-based models achieve better digit recognition and structure, particularly when template priors are included. Removing templates leads to reduced recognition and logic satisfaction, demonstrating their importance for early guidance. Interestingly, even the \texttt{weak\_ltn} configuration matches or exceeds other variants in digit classification, highlighting the robustness of symbolic supervision when paired with data-level diversity.

\begin{table}[h]
    \centering
    \resizebox{\textwidth}{!}{
    \begin{tabular}{l c c c c c}
        \toprule
        \textbf{Variant} & \textbf{Quality Score} & \textbf{Coverage} & \textbf{Digit Recognition} & \textbf{Template Dependence} & \textbf{Logic Satisfaction} \\
        \midrule
        baseline\_gan         & 0.360 & 0.069 & 0.717 & 0.700 & 0.000 \\
        full\_ltn\_gan        & 0.395 & 0.070 & 0.725 & 0.700 & \textbf{0.978} \\
        no\_ltn\_constraints  & 0.392 & 0.069 & 0.720 & 0.700 & 0.000 \\
        no\_templates         & 0.367 & 0.057 & 0.685 & 0.000 & 0.977 \\
        weak\_ltn             & 0.391 & 0.069 & \textbf{0.731} & 0.700 & 0.977 \\
        strong\_ltn           & \textbf{0.398} & 0.069 & 0.727 & 0.700 & 0.978 \\
        \bottomrule
    \end{tabular}
    }
    \caption{MNIST ablation study: logic boosts recognition and semantic accuracy.}
    \label{tab:mnist-ablation}
\end{table}

\subsection{Understanding Output Variability and the Impact of Logic Design}

Our ablation studies across all datasets reveal several important factors that influence the variability observed in LTN-GAN’s performance. While the integration of logic clearly improves structure, semantics, and generalization, the degree of improvement depends significantly on how the logical components are designed and implemented.

One key source of variability comes from how the logical constraints are formulated. For example, in the Ring dataset, we find that using overly simple constraints results in poorer geometric structure and reduced balance across sample categories. On the other hand, hierarchical and semantically rich logic leads to better ring adherence and improved logical satisfaction. This demonstrates that meaningful constraint design is essential for generating structured samples, especially in tasks where geometry or symbolic relationships matter.

Another important factor is the way logic is applied over the course of training. Variants that use progressive scheduling — gradually increasing the influence of logic over time — consistently outperform those where constraints are applied uniformly or abruptly. This staged approach allows the model to first explore the data space and later refine outputs according to logical structure. The benefit of such curriculum-like training is especially clear in the Ring and Grid datasets.

Constraint weighting also plays a major role. When all constraints are treated equally or fixed in importance, the model tends to either ignore weaker constraints or overfit to strong ones. Our experiments show that dynamic, hierarchical weighting — where constraints are emphasized based on their current satisfaction levels — helps balance between learning general patterns and maintaining logical structure. This strategy leads to better outcomes in both quality and logic adherence.

Additionally, we observe that different datasets respond differently to the same architectural changes. MNIST, for instance, shows stable performance across most ablations, likely due to its well-defined visual structure. In contrast, the synthetic datasets, which require explicit spatial relationships, are much more sensitive to constraint design and scheduling. This highlights the importance of dataset-specific logic design when applying LTN-based generative models.

Finally, our findings confirm a trade-off between sample quality and logic satisfaction. Increasing logical enforcement often leads to reduced sample variability, particularly in strongly constrained variants. This suggests that logic must be carefully balanced with generative flexibility to avoid mode collapse or overly rigid outputs.While LTN-GAN offers significant advantages over traditional GANs, achieving the highest performance requires careful tuning of constraint structure, logic weight scheduling, and dataset-specific considerations. Our results suggest that these models benefit from adaptive strategies rather than fixed configurations, particularly in domains where structural accuracy and semantic validity are essential.

\section{Discussion}

Our study demonstrates that LTN-GAN effectively integrates symbolic reasoning into the generative process, enabling the generator to produce outputs that not only resemble real data but also conform to domain-specific logical constraints. Across experiments, the model consistently outperformed unconstrained baselines in logic satisfaction and structural quality. However, this integration introduces a well-known tension: enforcing logical consistency often restricts sample diversity, limiting the generator's ability to explore the full data manifold. This trade-off was particularly evident in ablation studies, where higher constraint weights improved logical adherence but occasionally reduced visual variability or coverage. The generator’s objective blends adversarial and logical losses using tunable weights \(\alpha\), \(\lambda(e)\), and \(\beta\). These hyperparameters play a critical role in balancing quality, diversity, and rule satisfaction. While our scheduling strategies and architecture-specific adjustments allowed reasonable control, the need for careful tuning indicates that LTN-GAN still lacks adaptive mechanisms for managing this trade-off autonomously.

To situate LTN-GAN within the broader neuro-symbolic landscape, we note that several models embed symbolic structures in latent space to enforce rules indirectly. For example, VAEL \cite{misino2022vael} incorporates logic programming into VAEs, while MultiplexNet \cite{hoernle2022multiplexnet} models hierarchical categorical variables to guide generation. These approaches, like LTN-GAN, aim to impose structure, but differ by relying on probabilistic embeddings rather than direct logic reasoning within training objectives.

Other strategies approximate symbolic constraints with differentiable surrogates, making them compatible with gradient-based learning. Semantic Loss \cite{xu2018semantic} was one of the first to express logical rules as differentiable penalties. This has inspired methods like Pseudo-Semantic Loss \cite{ahmed2023pseudo}, which localizes constraint enforcement in autoregressive models, and Semantic Objective Functions \cite{mendez2024semantic}, which embed logic into loss functions using tools from information geometry. These approaches are powerful but often not tightly coupled with adversarial dynamics, where unstable gradients and non-stationary objectives complicate constraint enforcement.

Post-hoc constraint satisfaction through projection or refinement is another promising direction. For instance, SketchGen \cite{para2021sketchgen} and Vitruvion \cite{seff2022vitruvion} enforce geometric consistency in CAD sketches, while Langevin-based refinement \cite{kumar2022constrained} iteratively nudges generated outputs toward feasible regions. These methods can boost constraint satisfaction without affecting the core generative process, but they sacrifice training-time reasoning and may result in inefficiencies or brittle outputs.

Our findings also highlight the limitations of hard vs.\ soft constraint enforcement. While hard constraints ensure compliance, they can restrict diversity and exacerbate mode collapse. In contrast, soft penalties offer flexibility but often lack guarantees. Hybrid mechanisms have emerged to bridge this gap—e.g., Straight-Through Estimators for discrete constraints \cite{yang2022injecting}, and dual optimization methods that balance feasibility and exploration \cite{ferber2024genco, liu2020chance}. LTN-GAN adopts a soft logic-loss formulation, but further refinement may be needed to avoid local minima or underutilized rules.

One of the central gaps in current neuro-symbolic GANs is the lack of deep integration between symbolic reasoning and adversarial learning. Most models inject logic indirectly—through latent space priors or post-hoc corrections—rather than shaping generator updates directly. For instance, works like \cite{PengGenUnderstand23} rely on rule-injected representations but do not adapt rules dynamically. LTN-GAN takes a step forward by making logic satisfaction differentiable and jointly trainable, but it still relies on predefined rule sets. Future extensions may benefit from dynamic rule induction, where constraints are learned from data or inferred contextually. Meta-abduction \cite{PengGenUnderstand23} and latent-symbolic alignment \cite{CornelioWhenWhere23} offer promising foundations for this direction, while advances in goal-directed generation of structured discrete data \cite{mollaysa2020goal} demonstrate complementary strategies for enforcing properties directly through learned reward optimization.

Scalability also remains an open challenge. While LTN-GAN performs well on structured datasets like MNIST and synthetic geometries, scaling to higher-dimensional tasks (e.g., natural images or molecular graphs) is non-trivial. One direction involves hierarchical or geometric solvers \cite{seff2022vitruvion}, which may enable logic to operate at multiple levels of abstraction. Additionally, integrating constraint-aware discriminators—capable of assessing rule violations during training—could improve logic enforcement without sacrificing diversity.

Overall, our work contributes to the emerging field of constraint-aware generative modeling, aligning with efforts to combine logic and learning \cite{di2020efficient, StoianDCLG24, XueEmbedding19, hu2018deep}. Recent frameworks such as Generative Neurosymbolic Machines \cite{JiangGenNeSyMachine20} and abductive refinement methods \cite{PengGenUnderstand23} reinforce the potential of symbolic reasoning to enhance generative control. However, robust integration of logic within adversarial training remains underdeveloped. Addressing key challenges in constraint optimization, adversarial stability, and dynamic logic learning will be essential for realizing the full potential of neuro-symbolic GANs.

\subsection{Broader Implications for Generative Reasoning}

While the current study evaluates LTN-GAN within the standard generative adversarial framework, it is important to acknowledge that the potential of Logic Tensor Networks extends beyond improving sample quality or diversity in such settings. The constrained and adversarial nature of GANs often limits the full expressive power of symbolic logic integration. In particular, the generator is still primarily optimized for adversarial objectives, which may conflict with deeper forms of structured reasoning.

We believe that the true strength of LTNs lies in their ability to support logic-guided generative reasoning. This is especially relevant for tasks like scene generation, compositional image synthesis, or multi-modal symbolic grounding—areas where the generation process involves satisfying complex relationships, spatial configurations, and rule-based dependencies. In such contexts, LTNs can act not just as a regularizer but as a driver of structured generation, enforcing high-level semantic rules during the generative process itself. Future work should explore this direction by integrating LTNs into architectures beyond GANs, such as diffusion models, autoregressive decoders, or neuro-symbolic scene simulators, where symbolic reasoning can directly shape content generation rather than merely constrain it post hoc.

\section{Conclusion}

LTN-GAN represents a step forward in neuro-symbolic generative modeling, demonstrating that logical constraints can effectively guide adversarial learning. However, challenges such as the diversity-adherence trade-off, manual rule specification, and scalability constraints highlight key areas for future improvement. Research on adaptive rule learning, constraint-aware adversarial training, and scalable logic enforcement techniques will be crucial for advancing neuro-symbolic GANs into more robust and flexible frameworks. Addressing these challenges will enable GANs to generate logically structured, high-quality outputs across diverse domains while retaining generative diversity and flexibility.

\section{Acknowledgments}
Vaishak Belle was funded by a CISCO grant and a Royal Society University Research Fellowship.

\bibliographystyle{eptcs}
\bibliography{references}

\newpage
\appendix

\section{Extended Methodology}
\subsection{Detailed Experimental Setup}
\label{experimentalsetup}

We evaluate LTN-GAN against a baseline GAN that shares the same generator and discriminator architectures, optimization settings, and training schedule, but omits any logic-based supervision. This parallel setup ensures that observed performance differences stem from the integration of logical constraints, rather than incidental differences in architecture or learning dynamics.

Experiments are conducted across three synthetic datasets—\textit{Gaussian}, \textit{Grid}, and \textit{Rings}—as well as the image-based \textit{MNIST} dataset. In all cases, the generator is a fully connected multilayer perceptron (MLP) with LeakyReLU activations and hidden layers of size 128. For the synthetic datasets, the generator receives a 2D latent vector from a standard normal distribution and produces 2D samples with no output activation, enabling unbounded outputs in $\mathbb{R}^2$. The MNIST generator is conditioned on a one-hot digit label concatenated with 100-dimensional Gaussian noise and outputs $28 \times 28$ grayscale images. The discriminator matches the generator in structure, outputting a single probability score, and, for MNIST, an auxiliary classification head is added for digit prediction.

Training is performed using the Adam optimizer with learning rate $0.001$, $\beta_1 = 0.5$, and $\beta_2 = 0.999$, over 100 epochs. A batch size of 32 is used for synthetic data and 64 for MNIST. Discriminator label smoothing is employed: real samples are labeled with 0.9, and generated samples with 0.1. In all LTN-GAN settings, the generator loss combines adversarial and logical components: $L_G = \alpha L^{\mathrm{adv}} + \lambda(e) L^{\mathrm{logic}} + \beta L^{\mathrm{aux}}$, with $\alpha = 1$, $\beta = 0.2$ (if applicable), and $\lambda(e)$ controlled by a linear ramp-up schedule over the first $K$ epochs. For the Gaussian dataset, $\lambda$ rises from 0.05 to 0.3 over 80 epochs; for Rings and Grid, from 0.1 to 0.5 over 50 epochs; and for MNIST, from 0.05 to 0.25 over 60 epochs.

Additional stabilizing mechanisms are used in logic-augmented training. For Rings and MNIST, a checkpoint-based backtracking mechanism allows the generator to roll back to one of the previous 20 saved states if a sudden drop in rule satisfaction is detected. MNIST further incorporates a logic-aware curriculum: early training mixes in digit template priors for structure guidance, which are gradually faded out, letting the generator learn symbolic constraints autonomously. In all synthetic datasets, the generator receives added Gaussian noise (standard deviation 0.1) in the latent space to promote exploratory diversity. Evaluation is performed every 10 epochs by sampling 1,000 points from the generator and computing two primary metrics: (i) logical rule satisfaction using fuzzy logic aggregation over the constraint set, and (ii) geometric diversity, measured by pairwise Euclidean distances (or pixel variance for MNIST). MNIST additionally tracks classification accuracy and brightness/centering statistics through learned predicate networks. All experiments are implemented in PyTorch and executed on a MacBook Pro (2.4\,GHz 8-Core Intel Core i9, 64\,GB RAM, AMD Radeon Pro 5500M 8\,GB GPU).

To facilitate symbolic supervision, each dataset uses a custom set of logical predicates tailored to its geometric or semantic structure. For the synthetic domains, these predicates are implemented as fixed mathematical functions or simple threshold-based heuristics over coordinate space. In the Grid and Rings datasets, existential and universal constraints ensure spatial coverage and separation of geometric regions, with satisfaction computed via fuzzy logic aggregation. MNIST predicates, in contrast, are learned via small MLPs and are trained in tandem with the generator to evaluate properties such as digit class, visual clarity, and spatial alignment. Quantifiers are realized using differentiable power means—min-like for universal rules and max-like for existential ones—allowing gradients to flow through logical formulas. Logic satisfaction is tracked per rule, and adaptive weights are optionally applied to focus learning on underperforming constraints. This modular logic evaluation framework enables LTN-GAN to flexibly incorporate symbolic rules without disrupting the adversarial learning loop.

\newpage
\newpage
\begin{algorithm}[H]
\footnotesize
\caption{LTN--GAN Training Algorithm\protect\footnotemark[1]\protect\footnotemark[2]}
\label{alg:ltnganalgorithm}
\begin{algorithmic}[1]
\Require Training data $\mathcal{D}$, Generator $G$, Discriminator $D$, Constraint system $\mathcal{C}$ with rules $R$, config $\theta$ (incl.\ $\lambda(e)$), optional auxiliary loss $\mathcal{A}$
\Ensure Trained models $G^*, D^*$

\State Initialize parameters of $G$, $D$, and optionally $\mathcal{C}$
\State Create optimizers: $\mathrm{opt}_G$, $\mathrm{opt}_D$ (and $\mathrm{opt}_\mathcal{C}$ if applicable)
\State Set label smoothing values: $y_{\mathrm{real}} \leftarrow 0.9$, $y_{\mathrm{fake}} \leftarrow 0.1$
\State Define logic loss weight schedule $\lambda(e)$

\For{$e = 1$ \textbf{to} $N_{\mathrm{epochs}}$}
  \State $\lambda \gets \lambda(e)$
  \For{batch $B \subset \mathcal{D}$}
    \State Sample real data $x_R \sim B$, latent noise $z \sim \mathcal{N}(0, I)$
    \State Generate fake data $x_G \gets G(z)$

    \Comment{--- Discriminator Update ---}
    \State $\mathrm{opt}_D.\mathrm{zero\_grad}()$
    \State $L_D^{\mathrm{real}} \gets \ell_{\mathrm{BCE}}(D(x_R), y_{\mathrm{real}})$
    \State $L_D^{\mathrm{fake}} \gets \ell_{\mathrm{BCE}}(D(\mathrm{detach}(x_G)), y_{\mathrm{fake}})$
    \State $L_D \gets L_D^{\mathrm{real}} + L_D^{\mathrm{fake}}$
    \State \textbf{backprop} $L_D$, \quad $\mathrm{opt}_D.\mathrm{step}()$

    \Comment{--- Logic Constraint Evaluation ---}
    \State $\text{cons} \gets \mathcal{C}.\mathrm{evaluate}(x_G, R)$
    \State $(S_{\mathrm{logic}}, L_{\mathrm{constraint}}) \gets \mathcal{C}.\mathrm{aggregate}(\text{cons})$
    \State $\mathcal{C}.\mathrm{update\_weights\_or\_phase}(\text{cons})$

    \Comment{--- Generator Update ---}
    \State $\mathrm{opt}_G.\mathrm{zero\_grad}()$
    \If{$\mathcal{C}$ has learnable parameters}
      \State $\mathrm{opt}_\mathcal{C}.\mathrm{zero\_grad}()$
    \EndIf
    \State $L_G^{\mathrm{adv}} \gets \ell_{\mathrm{BCE}}(D(x_G), y_{\mathrm{real}})$
    \State $L_{\mathrm{aux}} \gets \mathcal{A}(x_G)$ \Comment{Optional, if $\mathcal{A}$ is defined}
    \State $L_G \gets \alpha L_G^{\mathrm{adv}} + \lambda L_{\mathrm{constraint}} + \beta L_{\mathrm{aux}}$
    \State \textbf{backprop} $L_G$, \quad $\mathrm{opt}_G.\mathrm{step}()$
    \If{$\mathcal{C}$ has learnable parameters}
      \State $\mathrm{opt}_\mathcal{C}.\mathrm{step}()$
    \EndIf

    \State Log metrics: $L_G$, $L_G^{\mathrm{adv}}$, $L_D$, $S_{\mathrm{logic}}$, $\lambda$, $L_{\mathrm{constraint}}$, discriminator accuracy
  \EndFor
\EndFor
\State \Return $G^*, D^*$
\end{algorithmic}
\end{algorithm}

\footnotetext[1]{%
\textbf{Loss Definitions.} Let $\ell_{\mathrm{BCE}}$ denote the binary cross-entropy loss. Discriminator losses:  
$L_D^{\mathrm{real}} = \mathbb{E}_{x_R}[\ell_{\mathrm{BCE}}(D(x_R), y_{\mathrm{real}})]$,  
$L_D^{\mathrm{fake}} = \mathbb{E}_{z}[\ell_{\mathrm{BCE}}(D(G(z)), y_{\mathrm{fake}})]$,  
with total $L_D = L_D^{\mathrm{real}} + L_D^{\mathrm{fake}}$.

Generator losses include:  
$L_G^{\mathrm{adv}} = \mathbb{E}_{z}[\ell_{\mathrm{BCE}}(D(G(z)), y_{\mathrm{real}})]$,  
$L_{\mathrm{constraint}} = 1 - S_{\mathrm{logic}}$ (or adaptive weighted sum),  
and optional $L_{\mathrm{aux}}$ for classification, template matching, or other tasks.

The overall generator loss is  
$L_G = \alpha L_G^{\mathrm{adv}} + \lambda(e) L_{\mathrm{constraint}} + \beta L_{\mathrm{aux}}$,  
where $\alpha$, $\beta$, and $\lambda(e)$ modulate the loss components.

\textbf{Logic Weight Schedule.} $\lambda(e)$ increases linearly over the first $K$ epochs:
\[
\lambda(e) =
\begin{cases}
\lambda_{\mathrm{start}} + \dfrac{e-1}{K-1}(\lambda_{\mathrm{end}} - \lambda_{\mathrm{start}}), & e \le K \\
\lambda_{\mathrm{end}}, & \text{otherwise}
\end{cases}
\]
}

\footnotetext[2]{\textbf{Dataset Notes:} Gaussian and Ring use linear $\lambda$ ramps. Grid logs both $L_{\mathrm{constraint}}$ and $\lambda L_{\mathrm{constraint}}$. Ring employs hierarchical and adaptive constraints. MNIST includes an auxiliary loss $L_{\mathrm{aux}}$ for classification and shape templates. No early stopping is used. All experiments compute $L_D$ as a plain sum (no $\tfrac{1}{2}$ scaling).}

\newpage
\section{Experimental Results and Visualizations}

\subsection{Gaussian Dataset}
\begin{figure}[h!]
  \centering
  \includegraphics[width=0.95\linewidth]{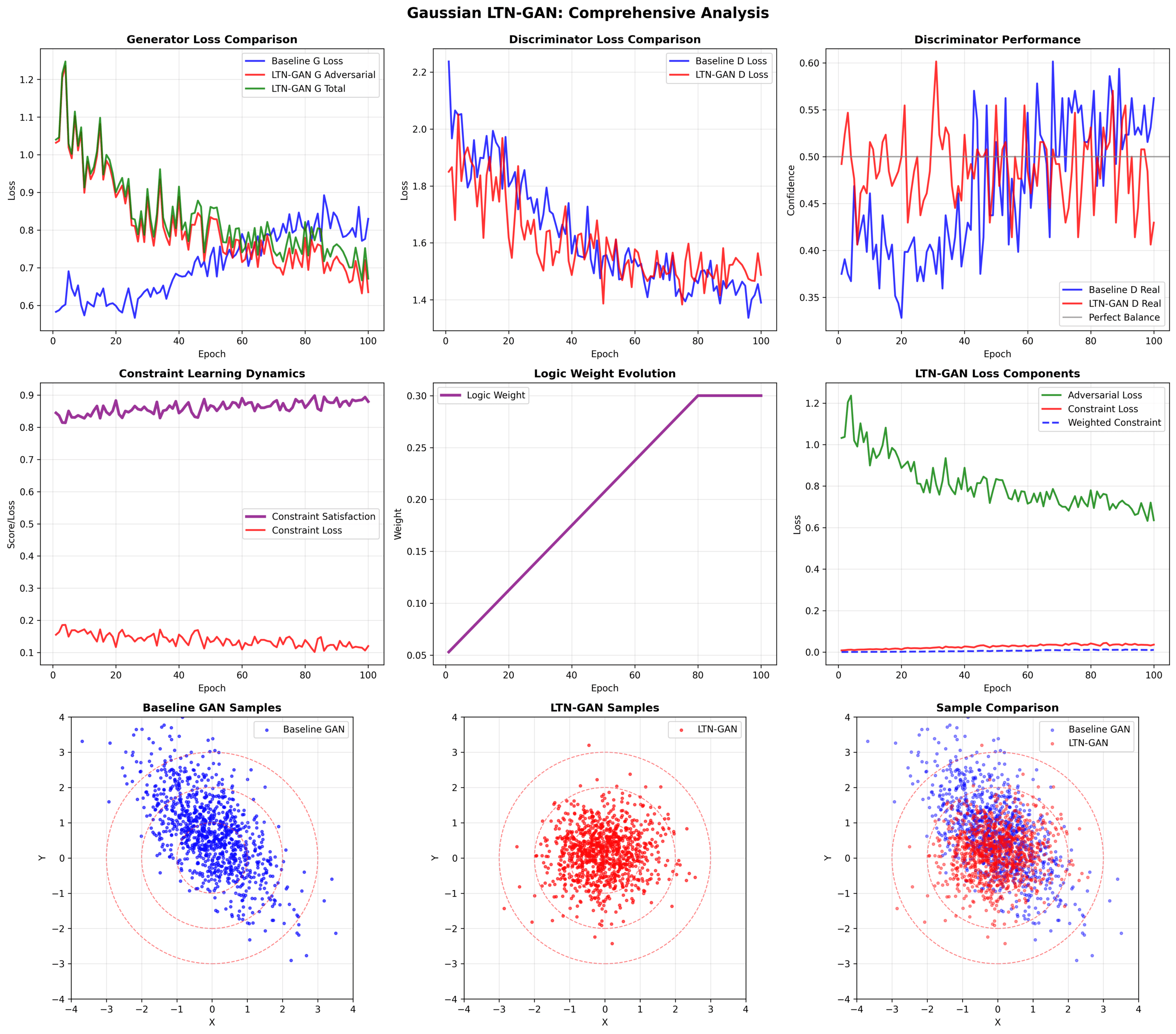}
  \label{fig:gaussian-results}
\end{figure}

For the Gaussian experiment, both predicates—\texttt{InRange}$(x)$ and \texttt{GaussianShape}$(x)$—are implemented as fixed mathematical functions rather than learned neural networks, enabling symbolic supervision without additional learning overhead. The generator is a fully connected MLP with LeakyReLU activations and no output activation, allowing it to generate samples across the full $\mathbb{R}^2$ plane. To stabilize training, label smoothing is applied (real samples labeled as 0.9, fake as 0.1), and the logic loss weight $\lambda(e)$ increases linearly from 0.05 to 0.3 over the first 80 epochs. Formula satisfaction is computed as the average truth degree of predicate conjunctions across the batch. Since all constraints are geometric and explicitly defined, the Gaussian task serves as a controlled testbed to isolate the effects of symbolic logic on GAN training without involving predicate learning.

\newpage
\subsection{Grid Dataset}
\begin{figure}[h!]
  \centering
  \includegraphics[width=0.95\linewidth]{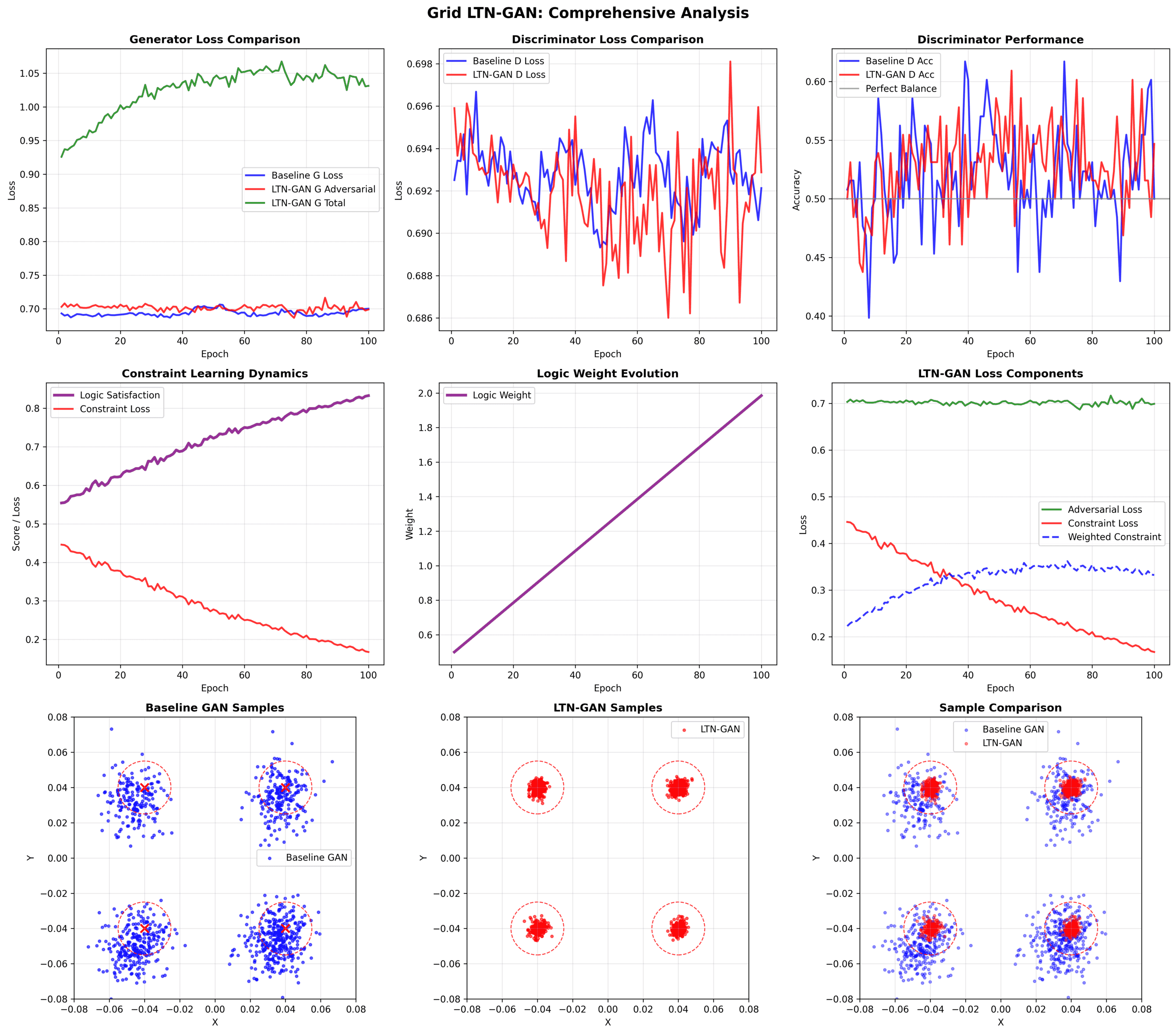}
  \label{fig:grid-results}
\end{figure}

In the Grid experiment, the generator is a fully connected MLP with LeakyReLU activations and no output activation, enabling unconstrained 2D outputs. A categorical selector network assigns each latent vector to one of four grid cells, and a perturbation is applied around that cell’s center to yield the final coordinate. This inductive bias guides initial structure formation. 

LTN predicates include a shared \texttt{OnGrid}$(x)$ network, evaluating proximity to any grid center, and four \texttt{InCell}$_i(x)$ networks, detecting cell-specific membership. These are trained jointly using distance-based features. Logical supervision is applied via a universal constraint over \texttt{OnGrid}$(x)$ and existential constraints over each \texttt{InCell}$_i(x)$, ensuring both alignment and coverage. The logic loss weight $\lambda(e)$ increases linearly from 0.5 to 2.0 over 100 epochs. Real samples are synthetic perturbations of grid centers (noise $\sigma=0.008$), and label smoothing (0.9/0.1) is used to stabilize training.

\newpage
\subsection{Rings Dataset}
\begin{figure}[h!]
  \centering
    \includegraphics[width=0.95\linewidth]{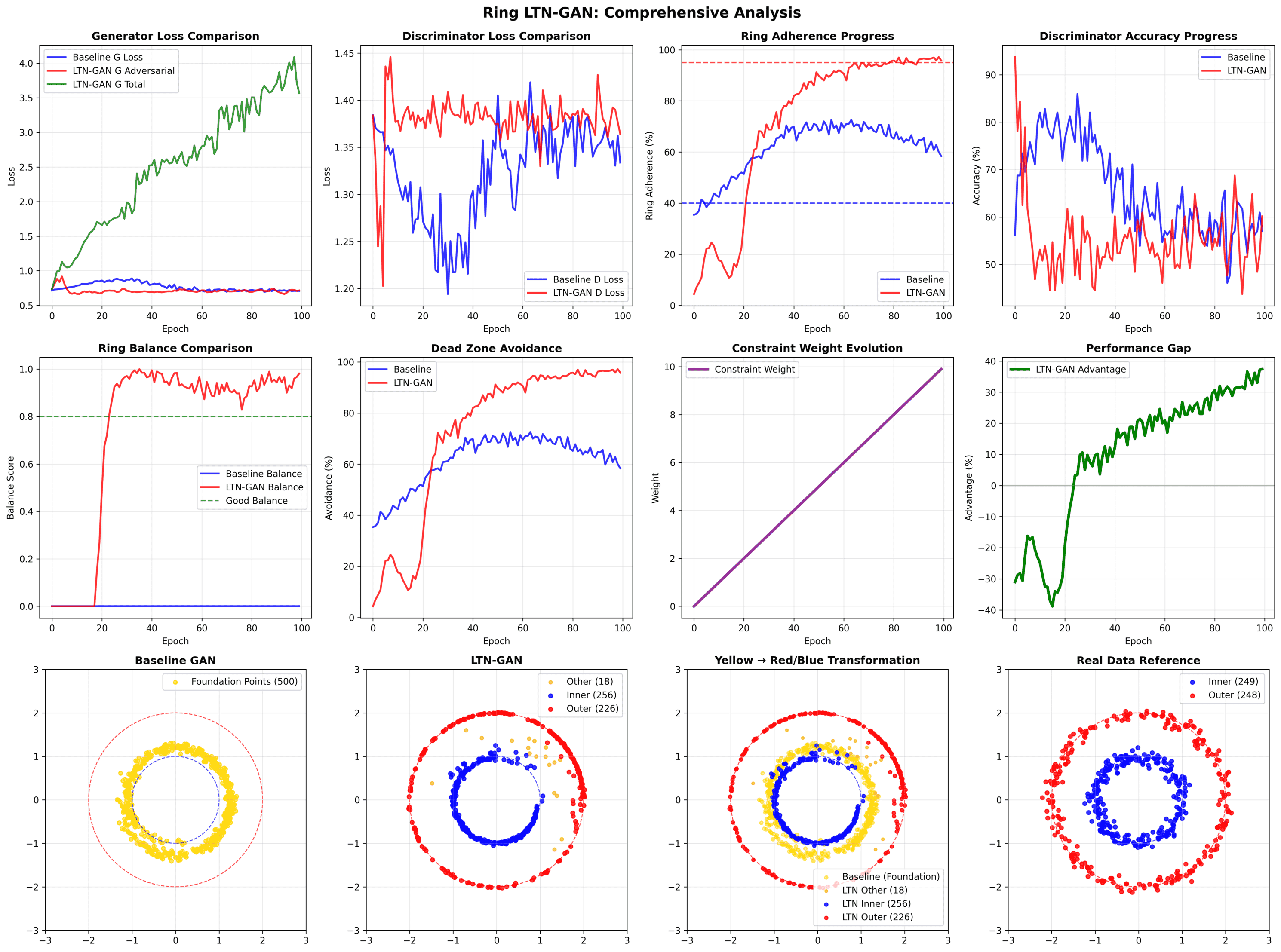}
    \label{fig:rings-results}
\end{figure}
In the Ring experiment, the data is sampled in polar coordinates and converted to Cartesian space, allowing for controlled radial distributions. Hardcoded geometric priors guide the generator to favor annular manifolds corresponding to inner and outer rings, while the dead zone in between is left unsampled. The generator is a multilayer perceptron with LeakyReLU activations and Tanh output to constrain the output space. Unlike in the Gaussian setting, most predicates—including \texttt{InnerRing}, \texttt{OuterRing}, \texttt{DeadZone}, and \texttt{NearCenter}—are implemented as differentiable neural classifiers trained jointly to provide flexible fuzzy memberships. A dynamic rule reweighting strategy updates clause weights based on per-batch satisfaction rates, allowing the model to adaptively prioritize under-satisfied constraints. The logic loss schedule begins at $\lambda=0.01$ and gradually increases to $\lambda=0.2$, with predicate tolerances (e.g., ring radius bands) tightened after warmup epochs to progressively refine precision.

\newpage
\subsection{MNIST Dataset}
\begin{figure}[h!]
  \centering
    \includegraphics[width=0.95\linewidth]{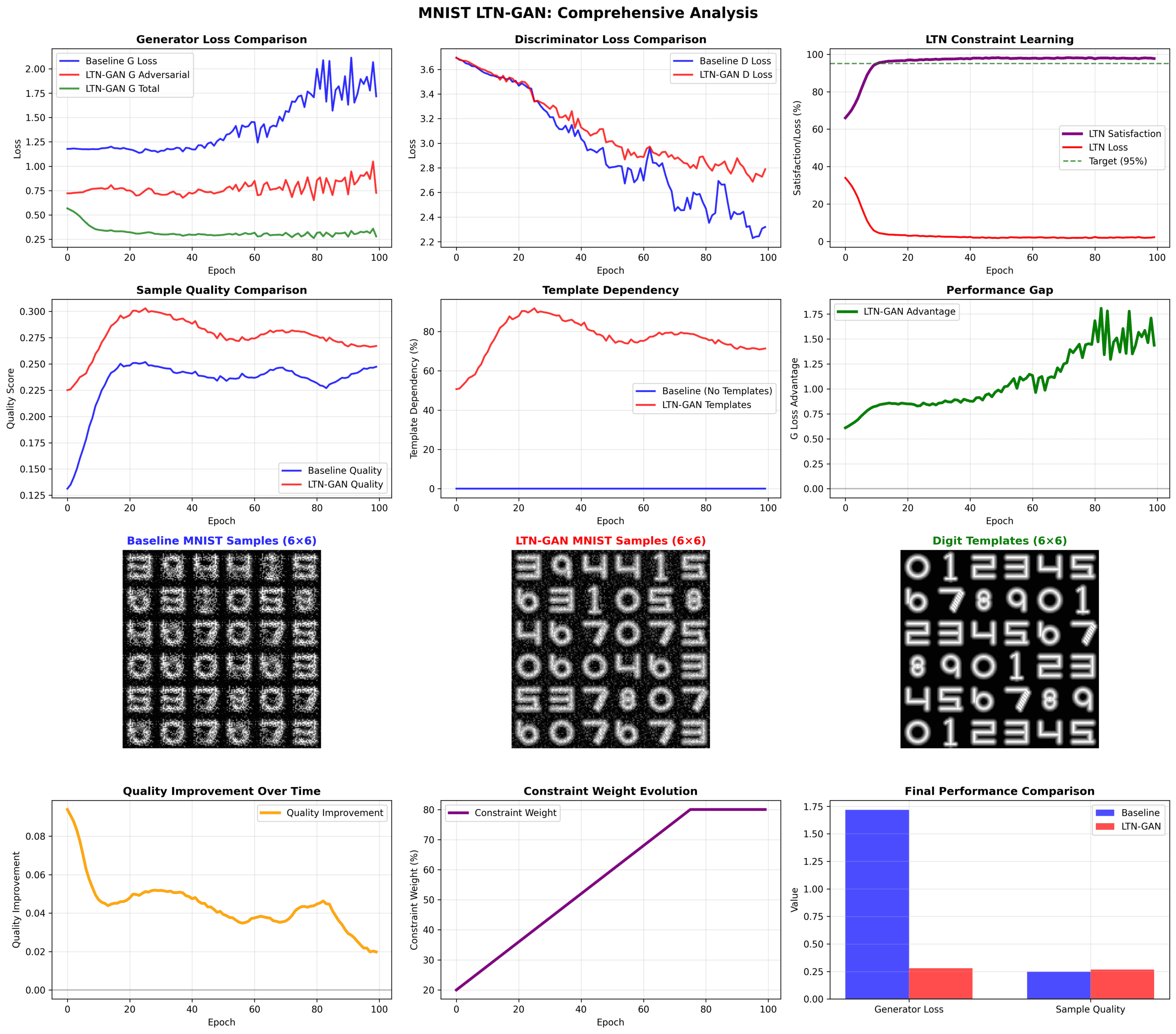}
    \label{fig:mnist-results}
\end{figure}

In the MNIST experiment, the generator is a convolutional neural network with transposed convolutions and batch normalization, trained to synthesize $28 \times 28$ grayscale digit images. Class predicates $\texttt{IsDigit}_k(x)$ for $k \in \{0,\dots,9\}$ are implemented as frozen MLPs trained on real MNIST digits, allowing symbolic supervision over digit identity. In addition, four structural quality predicates—\texttt{ValidPixels}$(x)$, \texttt{IsConnected}$(x)$, \texttt{IsComplete}$(x)$, and \texttt{HasProperIntensity}$(x)$—are defined using simple image processing heuristics and shallow neural nets. A handcrafted dataset of binary digit templates is used to blend geometric priors into the early training stages, guiding initial sample formation before symbolic constraints take effect. Logic loss weighting $\lambda(e)$ increases from 0.1 to 0.5 over the first 40 epochs. All predicates are evaluated on generated batches and their logic satisfaction scores are combined using the weighted fuzzy logic aggregation. This setup enables LTN-GAN to generate structurally valid and class-consistent images while preserving diversity across digits.
MNIST predicate networks are shallow CNNs trained end-to-end to capture digit properties such as centering and shape. Early training includes a blending of template priors, which are gradually phased out to shift supervision toward symbolic rules.

\newpage
\subsection{Domain-specific Metrics}

\begin{table}[h]
\centering
\scriptsize
\begin{tabular}{p{3cm} p{4cm} p{7cm}}
\toprule
\textbf{Dataset} & \textbf{Metric} & \textbf{Description / Computation} \\
\midrule

\multirow{6}{*}{Gaussian} 
    & Gaussian Adherence & Composite score based on stringent tests: normality (Anderson-Darling), bivariate shape (Mahalanobis distance), moments (mean $\approx 0$, std $\approx 1$), radial distribution (Rayleigh fit), independence (Pearson corr), and quartiles. Penalized if logic satisfaction is low. \\
    & Statistical Quality & Inverse of total deviation from target mean and std: $1 / (1 + \|\mu\| + \|\sigma - 1\|)$ \\
    & Combined Quality & Arithmetic mean of adherence and statistical quality. \\
    & Mean Error & L2 norm of difference between sample mean and $0$. \\
    & Std Error & L2 norm of difference between sample std and $1$. \\
    & Logic Satisfaction & Average truth degree of LTN constraints (range and Gaussianity). \\
    
\midrule

\multirow{6}{*}{Grid} 
    & Grid Cluster & Fraction of samples that belong to valid grid cluster centers (e.g., KMeans+thresholding on center match). \\
    & Coverage & Ratio of grid positions covered at least once. \\
    & Quality & Precision-like score evaluating how many points fall within valid tolerance of expected clusters. \\
    & Overall Performance & Weighted average of coverage and quality. \\
    & In Targets & Number of points within valid grid zones (rounded to integers or spatial masks). \\
    & Logic Satisfaction & Mean satisfaction across structural LTN constraints. \\

\midrule

\multirow{6}{*}{Ring} 
    & Ring Adherence & Fraction of samples lying near ring regions, measured via radial distance scoring and circular symmetry. \\
    & Inner / Outer Count & Number of samples classified inside or outside ring boundaries. \\
    & Balance Score & Normalized measure of symmetry between inner and outer zones (e.g., $1 - |n_{in} - n_{out}| / (n_{in} + n_{out})$). \\
    & Dead Zone Avoidance & Fraction of samples not falling into undesired central dead region. \\
    & Logic Satisfaction & Mean constraint satisfaction from LTN logic over spatial rings. \\

\midrule

\multirow{5}{*}{MNIST} 
    & Quality & Standard GAN image fidelity metrics (can include Inception-like proxy or visual diversity). \\
    & Coverage & Fraction of target digits generated (using classifier or template matching). \\
    & Digit Recognition & Accuracy of trained classifier over generated samples. \\
    & Template Dependence & Score measuring how many digits matched predefined templates. \\
    & Logic Satisfaction & Satisfaction over digit identity and position constraints defined in LTN logic. \\

\bottomrule
\end{tabular}
\caption{Overview of evaluation metrics used across dataset-specific ablation studies. Each metric is designed to align with both generative quality and logical structure fidelity.}
\end{table}

\end{document}